# A Double Deep Learning-based Solution for Efficient Event Data Coding and Classification

**ABDELRAHMAN SELEEM**[1, 2, 3]**, (Graduate Student Member, IEEE), ANDRÉ F. R. GUARDA**[2]**, (Member, IEEE), NUNO M. M. RODRIGUES**[2, 4]**, (Senior Member, IEEE), AND FERNANDO PEREIRA**[1, 2]**, (Fellow, IEEE)**

[1]Instituto Superior Técnico - Universidade de Lisboa, Lisbon, Portugal
[2]Instituto de Telecomunicações, Portugal
[3]Faculty of Computers and Information, South Valley University, Qena, Egypt
[4]ESTG, Politécnico de Leiria, Leiria, Portugal

Corresponding author: Abdelrahman Seleem (e-mail: a.seleem@lx.it.pt).

This work was funded by the Fundação para a Ciência e a Tecnologia (FCT, Portugal) through the research project PTDC/EEI-COM/1125/2021, entitled "Deep Learning-based Point Cloud Representation", and by FCT/MECI through national funds and when applicable co-funded EU funds under UID/50008: Instituto de Telecomunicações.

**ABSTRACT** Event cameras have the ability to capture asynchronous per-pixel brightness changes, usually called "events", offering advantages over traditional frame-based cameras for computer vision tasks. Efficiently coding event data is critical for practical transmission and storage, given the very significant number of events captured. This paper proposes a novel double deep learning-based solution for efficient event data coding and classification, using a point cloud-based representation for events. Moreover, since the conversions from events to point clouds and back to events are key steps in the proposed solution, novel tools are proposed and their impact is evaluated in terms of compression and classification performance. Experimental results show that it is possible to achieve a classification performance for decompressed events which is similar to the one for original events, even after applying a lossy point cloud codec, notably the recent deep learning-based JPEG Pleno Point Cloud Coding standard, with a clear rate reduction. Experimental results also demonstrate that events coded using the JPEG standard achieve better classification performance than those coded using the conventional lossy MPEG Geometry-based Point Cloud Coding standard for the same rate. Furthermore, the adoption of deep learning-based coding offers future high potential for performing computer vision tasks in the compressed domain, which allows skipping the decoding stage, thus mitigating the impact of compression artifacts.

## I. INTRODUCTION

In recent years, the fields of multimedia and computer vision (CV) have witnessed the growth of event camera-based systems, mainly due to the emergence of bio-inspired vision sensors. The innovative design of event cameras, exemplified by the Dynamic Vision Sensor (DVS) [1], represents a departure from conventional imaging acquisition devices. These sensors capture visual information depending solely on the measurement of the scene variation, asynchronously sampling the light based on the scene dynamics, as opposed to relying on a synchronous clock mechanism, unrelated to the observed scene [2]. Event cameras represent dynamic changes in brightness (i.e., relative changes above a preset threshold) using the so-called "events", which are independently and asynchronously generated for each pixel position. The output of an event camera is an asynchronous sequence of 4D events $(x, y, t, p)$, referred to as an *event sequence*, where $t$ corresponds to the timestamp, $(x, y)$ to the triggering pixel horizontal and vertical spatial coordinates, and $p$ to the polarity of the brightness change, i.e., increasing or decreasing [3].

This novel acquisition process allows to achieve high dynamic range, microsecond-scale temporal resolution, and energy-efficient operation. Furthermore, these sensors alleviate bandwidth requirements, making them well-suited for real-time, high-speed vision, and low power consumption [4]. The multiple advantages of event-based acquisition have led to its adoption in multiple applications, including object tracking [5], surveillance and monitoring [6], object/gesture recognition [7], depth estimation [8], structured light 3D scanning [9], image deblurring [10], star tracking [11], and real-time interaction systems [12]. Despite the operational advantages of event data, its wide use requires addressing two important challenges, notably the design of efficient coding and CV processing solutions.

Despite being the most natural solution, designing conventional CV tools directly operating on event data is challenging since event data is rather unstructured and has sparse characteristics, contrary to images. Deep learning (DL) has emerged as the dominant approach for exploiting the spatiotemporal information provided by event data for several CV tasks [13]-[17]. There are two main DL-based approaches for handling asynchronous and sparse event data: event-by-event processing and grid-based (or batch) processing [18]. The event-by-event approach directly handles the incoming events as they are received, minimizing the processing delay; however, training remains challenging and is still under development since a suitable loss function for backpropagation has not yet been established. In contrast, the grid-based approach preprocesses event data to convert it into a structured, grid-based format, thus enabling the use of Convolutional Neural Networks (CNN).

Classification is one of the most useful CV tasks, also for event data, and will be used in this paper as the target CV task. The DL-based Event Spike Tensor (EST) classifier [19] is a good example of a competitive DL-based event data classification solution; for this reason, it will be adopted in this paper and later presented in more detail. The EST classifier uses a grid-based approach known as "event spike representation" combined with a DL-based image classifier to classify event data, achieving gains in event data classification performance of approximately 12% over alternative relevant methods, notably classifiers using event frames [20], voxel grids [21] and HATS [14]. The EST classifier representation is versatile and can adapt to various event data tasks like optical flow estimation, achieving great flexibility and robustness across various scenarios [19]. Overall, these advantages position the EST classifier as a powerful tool for advancing the capabilities of event-based vision systems, particularly the classification task.

Event-based CV tasks are very often performed in practical application scenarios where the event data has to be previously stored or transmitted. Given the massive number of events in an event sequence, efficiently coding/compressing event data is essential for practical transmission and storage purposes. However, efficiently coding 4D event data is a challenge that still needs to receive significant research efforts; while there are a few works proposing lossless coding solutions, there is a notable lack of lossy coding solutions.

An emerging approach for coding event data involves converting the event sequence into a Point Cloud (PC) representation, thus enabling the use of available PC coding (PCC) solutions for event data coding [22]. A PC is defined as a collection of points in the 3D space with the $(x, y, z)$ coordinates representing the PC geometry. The $(x, y, t, p)$ event data sequence may be represented as a PC by considering $(x, y, z = t)$ as the spatial 3D coordinates to represent the PC geometry; to represent the polarity, $p$, either a single PC is used where the polarity is taken as an attribute of the $(x, y, z = t)$ geometry or two complementary PCs are created, one for each polarity. However, the different nature of the event PC third dimension (corresponding to the event data timestamp, $t$), notably in terms of precision/resolution, demands careful consideration when using this event data representation model.

The event to PC conversion approach facilitates the use of available PCC solutions, for example, the MPEG standards, notably Geometry-based PCC (G-PCC) for static PCC, and Video-based PCC (V-PCC) for dynamic PCC [22], which are based on conventional coding techniques. Motivated by the widespread success of DL across diverse visual-related research areas, JPEG has specified the JPEG Pleno Learning-based Point Cloud Coding standard adopting a DL-based approach. The current standard specification and associated reference software, from now on referred to as JPEG PCC, can outperform the G-PCC Octree and V-PCC Intra solutions, particularly for dense PCs [23]-[25]. The JPEG Pleno PCC standard scope targets learning-based representations offering not only efficient compression performance for human visualization but also effective CV tasks performance in the compressed domain [26]-[28] with significant complexity reductions.

While coding/compression is essential, it may be performed in a lossless or lossy way, with the latter introducing compression artifacts which may impact the effectiveness of CV tasks, if performed after decoding, such as in face recognition [29] and PC classification [30]. Regarding event data, most coding solutions in the literature adopt a lossless approach, notably using conventional coding tools, for example, G-PCC lossless [31]-[33], to avoid the impact of compression artifacts. However, lossy coding may achieve lower rates, eventually with negligible impact on the CV tasks performance. The move to event data lossy coding adopting a DL-based coding solution is especially interesting due to the high expectations and potential associated with event data compressed domain CV processing [26].

The objective of this paper is to propose a novel double DL solution for efficient event data coding and classification including lossy PC-based coding which is first designed and after extensively assessed. This approach represents a significant departure from conventional coding methods which mostly focus on lossless and conventional coding techniques. The key tools in the proposed framework are the DL-based JPEG PCC codec [23] and the DL-based EST event data classifier [19], which will be described in detail later in this paper. The proposed coding-classification solution includes novel Event to PC and PC to Event Conversion tools and will be assessed using the test-split of both the Neuromorphic-Caltech101 (N-Caltech101) [34] and the CIFAR10-DVS event datasets [35]. In this context, the main novel contributions of this paper are:

- Design of the first of its kind, double DL-based solution for efficient event data coding and classification;
- Design of novel, efficient event data to PC and PC to event data conversion tools;
- Design of a novel method for processing different polarity duplicates, which relies on the voxel occupancy probabilities estimated by the DL-based JPEG PCC;
- Adoption of a DL-based JPEG PCC lossy codec for event data coding, which achieves substantial reductions in data rate without negatively affecting the EST classifier performance;
- Performance assessment of various relevant event data classification pipelines, notably offering different trade-offs in terms of compression versus classification performances.

Furthermore, the proposed solution lays the groundwork for future advancements in developing compressed domain CV processing, notably classifiers, which can further mitigate the impact of compression artifacts and reduce the overall system complexity, as decoding would not be performed anymore since classification would happen directly over the compressed latents.

Experimental results show that the classification performance for JPEG PCC lossy coded event data surpasses the classification accuracy of G-PCC lossy at comparable rates and is very close to that of the original event data (and naturally also of G-PCC lossless coding). More precisely, JPEG PCC lossy coded and decoded event data classification performance offers around 65% gains when compared to G-PCC lossy coding, measured in terms of the Bjontegard-Delta Rate (BD-Rate) for the Top-1 and Top-5 performance metrics. These results demonstrate that it is possible to effectively use lossy event data coding in a way that does not compromise the classification performance (regarding lossless coding) while enabling considerable rate reductions and associated memory resources; this is a major proof of concept with a potential large impact in future event data coding research. Furthermore, results show that the scaling and voxelization tools used in the Event to PC Conversion module strongly influence the event data DL-based classification performance.

This paper is organized as follows: Section II briefly reviews the more relevant event data coding and classification solutions in the literature; Section III statistically characterizes the N-Caltech101 event dataset adopted as the main dataset to be later used for coding and classification targeting to better understand the data to be processed; Section IV describes the proposed double DL-based solution for efficient event data coding and classification along with the associated pipelines; Section V describes the adopted off-the-shelf event data coding and classification modules; Section VI proposes new solutions for the Event to PC and PC to Event Conversion modules; Section VII reports and discusses the performance assessment results; and finally Section VIII concludes the paper and identifies future research directions.

## II. RELATED WORK

This section offers a brief overview of related work on event data coding and classification. First, various coding solutions representing event data using different modalities are reviewed, notably methods transforming event data into video-like data to leverage the spatial and temporal correlations; later, the focus will be on coding solutions representing event data as PCs.

### A. EVENT DATA CODING

Event cameras generate asynchronous event data streams, which require specialized coding techniques. The unique properties of event data produced by DVS differ significantly from frame-based videos, making it challenging to directly apply conventional video coding methods. To address this challenge, a video coding approach, called Time Aggregation based Lossless Video Encoding for Neuromorphic Vision Sensor Data (TALVEN), has been proposed, which transforms the event stream into a video-like format with significant spatial and temporal correlation [36]. While TALVEN achieves superior compression ratios compared to alternative coding methods, it leads to a marginal decrease in temporal resolution due to the performed temporal event aggregation, which may limit its suitability for applications requiring very high time resolution.

Another coding strategy for event data leverages the spatial correlation by projecting events onto macro-cubes with $(x, y)$ coordinates representing the event sequence spatial information [37]. Then, two coding modes are used for coding each macro-cube, notably Address-Prior (AP) and Time-Prior (TP). In the AP mode, the coding includes the macro-cube spatial information $(x, y)$ and the event count at each position. For instance, if multiple events occur at a specific $(x, y)$ location, the macro-cube records the total event count for that location, and each event's timestamp is coded separately using delta coding. On the other hand, the TP mode organizes the events based on their timestamps, projecting them in increasing order. Events are projected relative to a central point, a spatial location with multiple events across neighboring pixels, which effectively compresses the spatial information. For both TA and TP modes, spatial and temporal coordinates residuals undergo encoding through a Context-Adaptive Binary Arithmetic Coding (CABAC)-based method. However, since these are separately encoded, the CABAC does not leverage the correlation between the spatial and temporal fields.

Recent advances recognize that event data may be represented as a PC in multiple ways [31]-[33]. The sparsity similarity between event data and PCs, coupled with proven spatial and temporal correlation in event data, motivates leveraging PCC solutions for event data coding. In [31], [32], event data is represented using two PCs, one for each polarity, losslessly coded using the G-PCC lossless standard. This PC-based approach targets analysis and compression efficiency, by separately exploiting the correlation in the positive (POS) and negative (NEG) polarity events.

In [31], the impact of time duration on PC lossless compression performance is investigated by dividing 60-second scenes into multiple PC slices, considering different slice durations, namely 1s, 5s, 10s, 20s, 30s, and 60s for each event sequence. Results indicate that larger PC slices lead to efficient compression performance using G-PCC lossless; however, a saturation point was observed, beyond which further increasing time duration yielded minimal benefits. Larger PC slices also introduce increasing coding delay and encoding complexity, emphasizing the need for a balanced approach targeting each specific use case. The study underscores the importance of considering trade-offs between compression efficiency, delay, and encoding complexity when determining the optimal PC size.

In [32], the conversion of event data to PC involves scaling the spatiotemporal coordinates using two different scales, notably $1 \times 10^3$ and $1 \times 10^6$ for the spatial and temporal coordinates, respectively. Additionally, two aggregation strategies are considered during the event to PC conversion, one based on a fixed number of event points and the other on a fixed temporal interval; after, lossless coding is applied. The results indicate that the strategy with a fixed number of event points tends to achieve better compression performance than the alternative strategy with a fixed temporal interval, notably since leading to equivalent spatial and temporal redundancies, what seems to be beneficial for PC compression.

While lossless coding is still the primary approach for event data coding, lossy coding solutions are emerging to offer different compression performance trade-offs. The impact of G-PCC lossy coding on event camera-based CV tasks is assessed in [33], notably by studying the impact of compression artifacts on various event-based vision tasks, including recognition, optical flow estimation, depth estimation, and video reconstruction. The study finds that high compression ratios may be achieved while maintaining acceptable performance for the event-based vision tasks, with the output quality depending on the compression ratio.

### B. EVENT DATA CLASSIFICATION

Several works focus on learning how to represent event data for classification purposes; however, many fail to make full use of all the available information in the input event sequence. For example, in [20], event data is represented as a two-dimensional event frame, but the temporal and polarity information are discarded. In [38], event data is depicted as a two-channel image without utilizing the timestamps. Likewise, [21] adopts a voxel-grid representation but neglects polarity. In [14], the Histogram of Time Surface uses a manually crafted event representation which overlooks temporal details. In contrast, the EST method [19] represents events using a grid-like format able to preserve the essential information. This event data representation can subsequently be processed with available image-based classifiers for event data classification. EST is the DL-based classifier adopted in this paper for event data classification, and will be described later in more detail.

### III. CALTECH101 EVENT DATASET ACQUISITION AND CHARACTERIZATION

After a brief description of the acquisition process, this section includes a characterization study of the main event dataset adopted in this paper, notably the N-Caltech101 event dataset [34], to better understand the features of the novel type of data to be coded and classified. This labeled event dataset was selected considering its wide adoption for event data classification [19]. For the event data characterization, this section proposes a set of meaningful metrics which are then applied to the N-Caltech101 event dataset. This characterization allows an in-depth analysis of the used dataset and offers important information to better understand the compression and classification performance presented later in this paper.

### A. N-CALTECH101 EVENT DATASET ACQUISITION

This subsection describes the acquisition process for the N-Caltech101 event dataset [34], which is an event-based version of the widely used Caltech101 image dataset; since it is a labelled dataset, it is commonly used for event data classification experiments [39]. Each original 2D image is used to generate a so-called event sequence, which includes a set of 4D $(x, y, t, p)$ events. Each event sequence in the Caltech101 event dataset has an associated classification label corresponding to one of various possible classes, notably animals, vehicles, and everyday objects.

The N-Caltech101 event dataset was generated using an actuated pan-tilt platform that moves a dynamic vision sensor while capturing the displayed (and fixed) Caltech101 images. Converting an image into an event sequence takes 300 ms and includes three saccadic movements, i.e., rapid movements that abruptly change the point of fixation, each lasting 100 ms. The movement intensity for the first and second saccades is similar, while the third saccade has slower and shorter movements, thus resulting in the production of fewer events. Prior to the acquisition process, each image was resized to ensure that its width and height do not exceed 240 and 180 pixels, respectively, to match the field of view of the Asynchronous Time-based Image Sensor (ATIS) used for recording the N-Caltech101 event dataset [34].

The canonical N-Caltech101 event dataset representation format uses 40 bits per event [34], structured as follows:

- Bit 0-22 (23 bits): Timestamp (in milliseconds);
- Bit 23 (1 bit): Polarity (0 for NEG, 1 for POS);
- Bit 24-31 (8 bits): $x$ position (in pixels);
- Bit 32-39 (8 bits): $y$ position (in pixels).

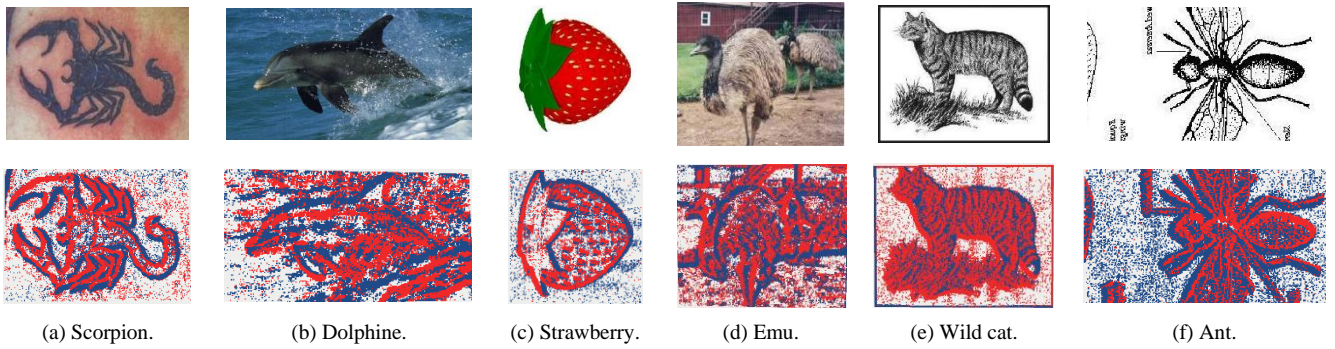

(a) Scorpion.    (b) Dolphine.    (c) Strawberry.    (d) Emu.    (e) Wild cat.    (f) Ant.

**FIGURE 1.** Samples from the Caltech101 image dataset (top) and the corresponding N-Caltech101 event dataset (bottom); each event sequence is represented using the superposition (in time) of all their POS (+) and NEG (-) events, depicted in blue and red, respectively.

The POS and NEG polarities are associated with brightness increases and reductions, respectively. Fig. 1 shows examples of images and the corresponding event sequences from the Caltech101 image and N-Caltech101 event datasets, respectively. The dataset includes 8,709 event sequences organized into 101 classes (4,356 for training, 2,612 for validation, and 1,741 for testing); each class comprises between 40 and 80 different event sequences. The test split of the N-Caltech101 event dataset, containing 1,741 event sequences, will be used for the characterization reported in the next subsections.

### B. EVENT CHARACTERIZATION METRICS

This subsection proposes a set of metrics for event dataset characterization. The metrics may be global or polarity-based, depending on whether they consider both POS and NEG polarities together without distinction or separately consider POS and NEG polarities. The proposed event data characterization metrics are:

- **Total number of events per event sequence** – Counts the number of events per event sequence and allows to analyze the distribution of the total number of events per event sequence; this metric can be used globally considering both event sequence polarities or considering each polarity independently.
- **Number of events along time** – Counts the number of events for bins along time and allows to analyze how the events are distributed along time; this metric can be used globally considering both event sequence polarities or considering each polarity independently.
- **NEG-POS events ratio per event sequence** – Measures the NEG-POS ratio per event sequence and allows to analyze the balance between the number of NEG and POS events per event sequence.
- **Event sequence sparsity** – Measures sparsity as the median of the mean distances between each event and its $k$ nearest neighbors and may be formally expressed as:

$$Sparsity = median\{d_{vmE}\}, \qquad (1)$$

where:

$$d_{vmE} = \frac{1}{k}\sum_{i=1}^{k}\|ev - ev_k\|^2, \forall\, ev \in V, \qquad (2)$$

with $d_{vmE}$ representing the mean Euclidean distance between each event, $ev$, in the event sequence $V$ and its $k$ nearest neighbors. This metric allows to analyze the sparsity of the distribution of events in the 3D $(x, y, t)$ space for each event sequence, as well as for each polarity independently.

- **Polarity coherence** – Measures the polarity coherence as the percentage of events for which at least $n$ of the $k$ nearest neighbor events share the same polarity as the current event and allows to analyze the coherence of the polarity for neighboring events in an event sequence.

While the metrics above are defined for each event sequence, the average ($\mu$) and standard deviation ($\sigma$) over the full dataset may be also computed to characterize the complete dataset.

The following subsection reports and analyzes the results for the defined global and polarity-level metrics applied to the test-split of the N-Caltech101 event dataset.

### C. N-Caltech101 Characterization and Analysis

This subsection reports and analyses the results obtained by applying the metrics proposed above to the 1,741 event sequences in the test-split of the N-Caltech101 event dataset. In this analysis, a Temporal Scaling Factor (TSF) of 256 was used to scale the original timestamp values for metrics that require computing distances, namely "event sequence sparsity" and "polarity coherence". This scaling is important to obtain a similar range of values for both the spatial and temporal coordinates. The specific value of 256 has been selected due to its relevance in terms of compression and classification performances, as will be shown in later sections.

Fig. 2 shows a graphical representation of the statistics for the proposed metrics, considering both the global and polarity-level variants, when relevant. The data distribution is shown using: i) density plots, representing the data probability density function; ii) box plots, indicating the median, quartiles, and data range; and iii) jitter plots, displaying the individual data points in a swarm plot. Furthermore, average values and

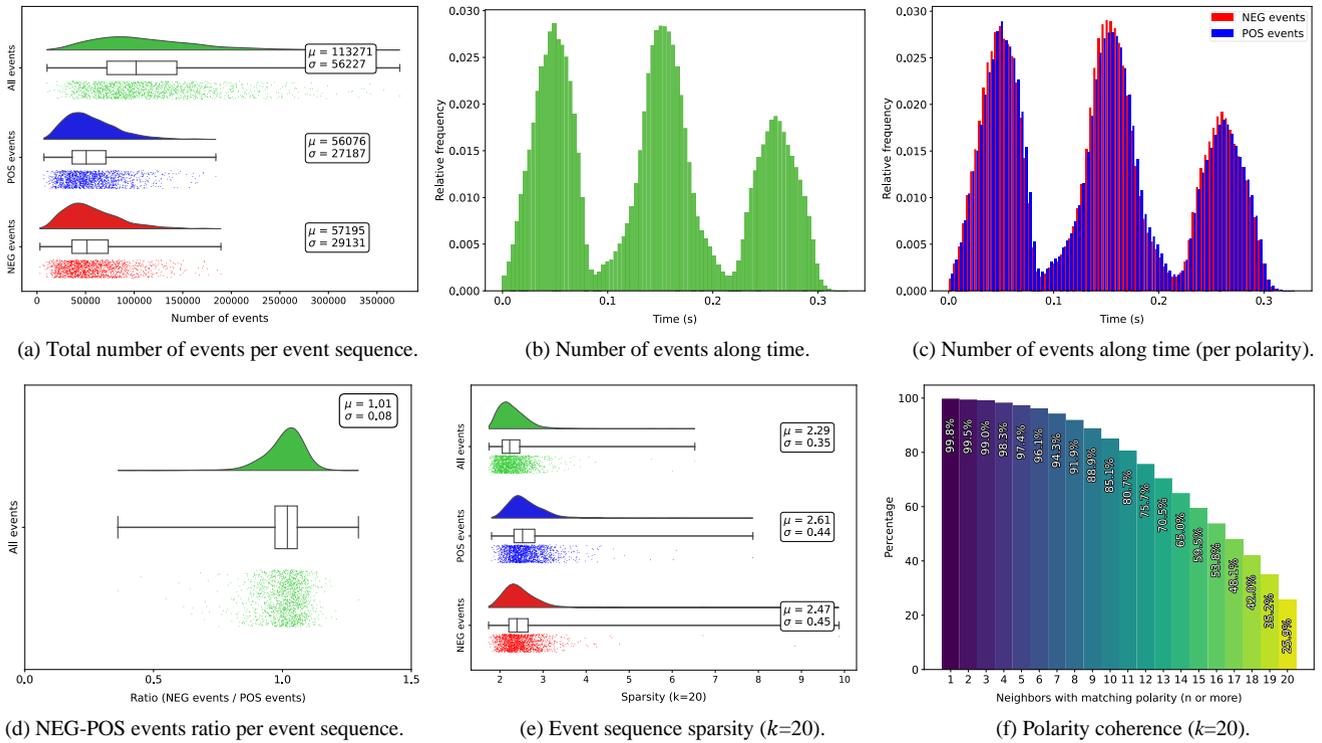

(a) Total number of events per event sequence.

(b) Number of events along time.

(c) Number of events along time (per polarity).

(d) NEG-POS events ratio per event sequence.

(e) Event sequence sparsity ($k$=20).

(f) Polarity coherence ($k$=20).

**FIGURE 2.** Statistics for the event characterization metrics applied to the test split of the N-Caltech101 event dataset.

standard deviations are included where relevant. From Fig. 2, it is possible to observe:

- **Total number of events per event sequence** – The distribution of the total number of events per event sequence is shown in Fig. 2a) as distribution, box plot and jitter plots. At global-level (i.e., considering both polarities), it is possible to observe a significant variability on the total number of events associated to each event sequence, ranging from 10,306 to 373,433 events with a mean of 113,271 and a standard deviation of 56,227. The same variability can be generally observed for the events of each polarity, with the expected adjustment in the mean and standard deviation values by a factor of approximately 0.5. At polarity-level, it is possible to see a slight imbalance between the distributions for the total number of POS and NEG polarities events, notably 99,577,067 (50.49%) and 97,628,147 (49.51%) for the total number of NEG and POS events, respectively. In fact, Fig. 2a) confirms that there are, on average, slightly more NEG ($\mu = 57,195$) than POS ($\mu = 56,076$) events, with a corresponding difference also on the standard deviation, i.e., 29,131 and 27,187 for NEG and POS events, respectively.

- **Number of events along time** - The distribution of the number of events along time is represented in the histograms of Fig. 2b) and Fig. 2c), for both polarities together and POS and NEG events separately, respectively. These histograms highlight the three distinct saccades performed during the data acquisition process. In both figures, it is possible to observe the uniform saccade duration and the higher number of events for the first two saccades. The distribution of POS and NEG events shown in Fig. 2c) is not exactly even over time. At the start and end of each saccade, the POS events tend to outnumber the NEG events while, in the middle of the saccades, NEG events outnumber POS events.

- **NEG-POS events ratio per event sequence** – The NEG-POS polarity ratio per event sequence shown in Fig. 2d) highlights that the distribution of NEG and POS events is slightly unbalanced. The overall average NEG-POS ratio for the test split of the N-Caltech 101 event dataset is 1.01 with a standard deviation of 0.08.

- **Event sequence sparsity** – The event sequence sparsity with a $k = 20$ neighborhood is shown in Fig. 2e). To perform a meaningful analysis of the event sequence sparsity, namely considering the coding goal of this paper, the sparsity has been computed after voxelizing the event data, i.e., rounding the (scaled) timestamp values to integers, and discarding the duplicate events. The average distance to the nearest 20 neighbors is $\mu$=2.29, with a standard deviation of $\sigma$=0.35, when considering events of both polarities. The relatively low standard deviation over the event sequences shows that they have a rather similar sparsity level. A slight increase on the $\mu$ and $\sigma$ values can be observed when both polarities are considered independently, with $\mu$ raising to 2.61 and 2.47 and $\sigma$ to 0.44 and 0.45, for POS and NEG events, respectively.

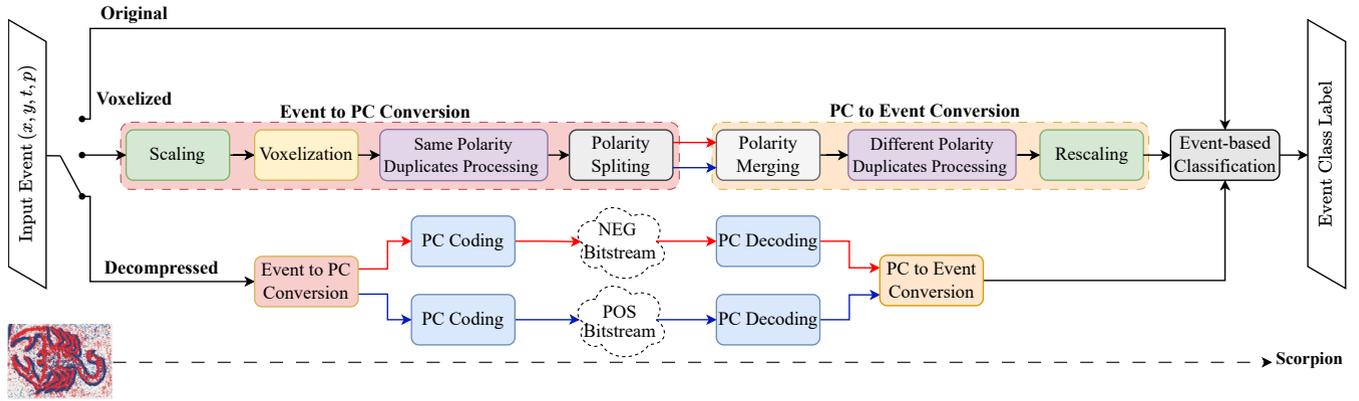

**FIGURE 3.** Double DL-based event data coding and classification architecture.

This indicates that, overall, the NEG events are slightly less sparse than the POS events maybe because there are slightly more NEG events.

- **Polarity coherence** – Despite the slight bias on the NEG-POS ratio, Fig. 2f) shows a high level of polarity coherence between neighboring events for this dataset. The results, for $k = 20$ and $n$ ranging from 1 to 20, clearly show that there is a very high probability that the events share the same polarity with their nearest neighbors. Specifically, 99.8% of the events share the same polarity with at least 1 of the nearest 20 neighbors, and 25.9% of the events share the same polarity with all 20 nearest neighbors; for $n$ equal to 2, 8, and 18, the percentages are 99.5%, 91.9%, and 42%, respectively, showing that the slope of the reduction increases with higher value of $n$.

It is relevant to highlight that a minor percentage (around 0.05%) of events in the test-split of the N-Caltech101 event dataset are duplicated events, meaning that some $(x, y, t)$ coordinates exist for both POS and NEG polarities. Due to their small proportion, these duplicated events were retained in the event dataset since they may have some physical meaning.

## IV. DOUBLE DL-BASED EVENT DATA CODING AND CLASSIFICATION: ARCHITECTURE AND PIPELINES

This section presents the architecture proposed for the double DL-based solution for efficient event data coding and classification framework, along with the associated classification pipelines.

### A. ARCHITECTURE AND KEY MODULES

Fig. 3 presents the overall architecture, including the three classification pipelines, applied to the input event sequence $(x, y, t, p)$. The classification pipelines include all or part of the following four major modules:

1. **Event to PC Conversion** – Converts the input event sequence $(x, y, t, p)$ into event PC data to enable event data coding with existing PCC solutions. Two distinct PCs, corresponding to the events with POS and NEG polarities, are generated in this conversion

(represented in Fig. 3 by the red and blue arrows, respectively). The Event to PC Conversion requires performing scaling and voxelization to create PCs with integer coordinates as needed for the PC Coding module.

2. **PC to Event Conversion** – Converts the event PCs of both polarities back to event data, by appropriately reversing the scaling operation performed in the previous Event to PC Conversion; it may be applied with or without PC Coding/Decoding modules in the pipeline.

3. **PC Coding** – Encodes the event PCs, in this paper using three selected PCC solutions: i) conventional G-PCC (lossless); ii) conventional G-PCC Octree (lossy); and iii) DL-based JPEG PCC (lossy). DL-based JPEG PCC is used for event data coding for the first time in the literature and it has a high potential for compressed domain processing where the decoding process is skipped.

4. **PC Decoding** – Decodes the event PCs bitstream using the appropriate decoder for the selected PCC solution.

5. **Event-based Classification** – Performs event sequence classification using a selected DL-based event data classifier; in this paper, the EST classifier was selected.

While Section V describes the selected off-the-shelf DL-based PCC and event classification solutions that are adopted in this paper, Section VI will present in detail the novel algorithms for the Event to PC and PC to Event Conversion modules.

### B. EVENT DATA PROCESSING PIPELINES

The architecture in Fig. 3 may be traversed according to three usage scenarios, thus creating three associated pipelines, before finally performing event data classification, each with a different impact on the final classification performance. The three pipelines represented in Fig. 3 from top to bottom are:

1. **Original Event Data Classification Pipeline** – This pipeline directly classifies the original floating-point precision event sequence $(x, y, t, p)$ using the selected

DL-based event data classifier. This pipeline generates the reference classification performance since no artifacts are introduced in the events that are input to the classifier.

2. **Voxelized Event Data Classification Pipeline** – This pipeline involves classifying input events that have been subjected to the Event to PC and the PC to Event Conversion processes which always includes scaling and voxelization, even if the pipeline name only refers 'voxelization'; there is no coding, and thus no compression artifacts, in this pipeline. This pipeline allows the evaluation of the impact of the scaling and lossy voxelization operations, required to produce the event PCs, on the final event data classification performance. Performing coding with a lossless solution like G-PCC lossless corresponds to using this pipeline since the lossless coding process introduces no compression artifacts.

3. **Decompressed Event Data Classification Pipeline** – This pipeline considers the classification of events that have been converted to an event PC representation and then subjected to lossy PC Coding/Decoding at a certain rate-distortion (RD) trade-off. The reconstructed event PCs are converted back to the original event data format and then passed to the selected DL-based event data classifier. This pipeline allows assessing the impact of lossy event data coding and its associated compression artifacts on the event data classification performance.

The three pipelines presented in Fig. 3 are used independently to demonstrate the impact of voxelization and lossy coding on event data classification. Each of the later two pipelines corresponds to a different, individual and, relevant processing path for the event data, which can significantly impact the final classification performance, e.g. with or without voxelization or lossy coding, when compared with the classification of the original event data, corresponding to the first pipeline.

The Event to PC Conversion solution adopted in this paper converts each event sequence into two separate PCs (one per polarity) for subsequent coding using PCC solutions, thus creating two bitstreams to be decoded. An alternative approach would be to concatenate the two polarity-based PCs into a single event PC, thus creating a single bitstream after PC Coding. In practice, both these approaches are expected to yield the same compression performance for the same coding solution. For simplicity, the proposed architecture graphically shows the event data coded as two separate PCs.

## V. DOUBLE DL-BASED EVENT DATA CODING AND CLASSIFICATION: BACKGROUND MODULES

This section briefly presents the two off-the-shelf selected key DL-based components for the proposed architecture: the DL-based PC codec (JPEG PCC) [23], and the DL-based event data classifier (Event Spike Tensor, EST) [19].

### A. DL-BASED JPEG PCC OVERVIEW

This subsection briefly describes the selected DL-based PC codec, notably the JPEG PCC codec [23]-[25].

Though JPEG PCC uses two different DL-based models to code the geometry and texture of a PC, this paper focuses solely on the JPEG PCC geometry coding part since the event data is converted only into PC geometry. JPEG PCC first divides the input PC into 3D blocks to facilitate random access and restrain the computational resources required for coding. Then, optionally, it down-samples the input PC block by a selected sampling factor, which is important for achieving lower coding rates or improving the compression efficiency for sparser PCs. The main JPEG PCC component is the DL coding model used to code the 3D blocks of binary voxels from the input PC, where each voxel can be either occupied (1) or empty (0). Following decoding, an up-sampling operation is applied to reverse the down-sampling process that may have been performed at the encoder. This optional up-sampling process may also involve a DL-based super-resolution model, which helps improve the quality of the reconstructed version of the coded PC block when down-sampling is performed.

The JPEG PCC coding model follows an autoencoder (AE) architecture with a variational hyperprior. The autoencoder first creates a compact, latent representation for the input PC block, which is then quantized and entropy coded using a probability distribution dynamically estimated with a variational autoencoder (VAE). At the decoder side, the latent representation is passed through the autoencoder decoding layers to generate the reconstructed (decompressed) PC block. This process performed by the AE decoding model outputs voxel occupancy probabilities which are subsequently binarized to determine the reconstructed points using a threshold optimized at the encoder.

The AE and VAE coding models are end-to-end trained for different target rates under the PCC Common Training and Test Conditions (CTTC) [40] set by JPEG, which defines a training dataset including 28 static PCs with different features, mainly regarding their density. The training process is controlled by a Lagrangian cost function, which uses a control parameter $\lambda$ to define a trade-off between rate and distortion/quality. The JPEG CTTC define five trained DL coding models with different $\lambda$ values, targeting to encode PCs at different RD points; these are the coding models used in this paper to obtain multiple RD points. For a more in-depth description of the JPEG PCC codec, refer to [23]-[25].

In terms of RD performance, JPEG PCC overcomes both G-PCC Octree and V-PCC Intra, particularly for coding dense PCs. For sparse PCs, the down/up-sampling modules are crucial to improve the compression performance, since they densify the input 3D block to provide the DL coding model a block more similar to those used for training.

## B. DL-BASED EVENT SPIKE TENSOR CLASSIFIER OVERVIEW

This subsection briefly describes the DL-based Event Spike Tensor (EST) [19] classifier, which has been selected due to its competitive classification performance, notably for the N-Caltech101 event dataset mainly adopted in this paper.

As previously stated, an event sequence consists of individual events that occur over time and are characterized by their spatial coordinates, timestamp, and polarity. To process this type of data, it is essential to convert the event data into a format that can be used by CNN architectures. Previous event data processing methods have used a preprocessing step to convert event data into a grid-like representation, which can be processed by traditional neural networks [19]. However, these methods have not considered the impact of the input representation on the downstream task performance. In contrast, EST [19] uses a novel event representation that is learned end-to-end together with the task itself, thus allowing the optimization of the event data representation for the specific task at hand, in this case, event data classification.

To classify an event sequence, the input event data, is transformed into an EST representation, which is a 4D data structure. In order to use a conventional image classifier, EST uses 9 bins for the temporal dimension while the polarity information is concatenated along the temporal dimension, generating an image-like representation with dimension $18 \times 224 \times 224$, where 18 is the number of channels with each channel corresponding to a $224 \times 224$ resolution image with a single component. Since ResNet-34 is used as the image classifier to predict the event sequence label, the first and last layers are adjusted to be retrained for the event classification task. Random initialization was used for training these two layers, followed by a fine-tuning step for all the ResNet-34 layers/weights.

The EST classifier has shown event data classification performance gains of approximately 12% [19] over relevant alternative methods, including event frames [20], voxel grids [21] and HATS [14].

## VI. EVENT TO PC AND PC TO EVENT CONVERSIONS

This section proposes novel solutions for the Event to PC and PC to Event Conversion modules, which are key components of the proposed double DL-based event data coding and classification architecture presented in Fig. 3.

### A. EVENT TO PC CONVERSION

This subsection proposes the Event to PC Conversion method to convert the input event sequence, represented by a sequence of $(x, y, t, p)$ values, into event PC data. The result of this conversion is a pair of complementary PCs, each composed by the points corresponding to the events of one given polarity (POS or NEG). This polarity-based PC splitting approach allows exploiting the high correlation observed among events of one given polarity.

It is proposed that the novel Event to PC Conversion method, represented in detail in Fig. 3, includes the following processing steps:

- **Scaling** – Scales the event sequence timestamps to produce a suitable event PC for coding purposes. Since the spatial coordinates and timestamp values have different ranges, scaling the temporal coordinate with a defined TSF is essential, while the same does not happen for the spatial coordinates. Thus, the scaled coordinates for the event PC $(x, y, z, p)$ are defined as:

$$(x, y, z, p) = (x, y, t.TSF, p) \qquad (3)$$

where $(x, y, t)$ are the spatial coordinates and timestamp values from the input event sequence.

- **Voxelization** – Performs the voxelization of the event data by rounding the scaled event PC coordinates to integers, commonly required for coding with voxel-based PCC solutions, such as G-PCC and JPEG PCC; this process affects particularly the temporal dimension, since the spatial coordinates are already voxelized.

- **Same Polarity Duplicates Processing** – Eliminates duplicated events, i.e., two or more events with the same spatial/temporal position and the same polarity, that may have been created with the voxelization process.

- **Polarity Splitting** – Splits the voxelized event data into two event PCs, one for each polarity, to be independently coded.

The choice of the TSF value has a major impact on the event PC characteristics, notably the PC sparsity, and the subsequent compression performance, as will be discussed in Section VII.

### B. PC TO EVENT CONVERSION

This subsection proposes a conversion method for the two single-polarity event PCs back to event sequence data. It is proposed that the novel PC to Event Conversion method, represented in detail in Fig. 3, includes the following processing steps:

- **Polarity Merging** – Combines the voxelized (if there is no coding) or reconstructed (if there is coding) event PCs from both polarities into a single voxelized or reconstructed event PC data, respectively. However, if lossy coding is used, there may be compression artifacts, notably the creation of duplicated points/events, i.e., points with the same spatiotemporal coordinates which are present in both the POS and NEG polarity event PCs.

- **Different Polarity Duplicates Processing** – Handles the different polarity duplicates, i.e., avoids that a single point may have both polarities as a consequence of the lossy coding process. To achieve this, the following methods are proposed:
  1. **Nearest neighbor polarity selection (NN)** – The polarity of each duplicate point/event is defined as the same as the nearest non-duplicate event; if there are two equally close events with opposing polarities, the algorithm selects the next nearest event until a majority decision is reached.

2. **Probability-based polarity selection (Prob)** – The polarity of each duplicate point/event is defined based on the voxel occupancy probabilities estimated by the AE decoding model; refer to [20] [21] for more details. This method selects the polarity of the corresponding voxel with the highest probability between the two produced by the AE decoding models associated to the NEG and POS polarities PCs. This polarity selection method is specific to JPEG PCC decompressed event PCs and can only be used in the Decompressed Event Data Classification pipeline. This is an advantage of DL-based PCC over conventional PCC solutions.

In the Voxelized Event Data Classification pipeline, the NN polarity selection method is used for processing different polarity duplicates (since there is no coding and thus no decoding probabilities available). For the Decompressed Event Data Classification pipeline, both the NN (for G-PCC lossy and JPEG PCC) and the Prob methods (only for JPEG PCC) are used.

- **Rescaling** – Rescales the voxelized/reconstructed event PC coordinates back to the range of values used in the original event sequence data. This rescaling uses the same TSF value that was used in the Event to PC Conversion module. After rescaling, the spatial coordinates and timestamps for the voxelized/reconstructed events are given by:

$$(\hat{x}, \hat{y}, \hat{t}, \hat{p}) = (\hat{x}, \hat{y}, \hat{z}/TSF, \hat{p}) \qquad (3)$$

where $(\hat{x}, \hat{y}, \hat{z}, \hat{p})$ represents the non-rescaled voxelized/reconstructed event PC data coordinates.

These voxelized and reconstructed event data correspond, respectively, to the voxelized and decompressed event data classification pipelines, described in Section IV.

## VII. PERFORMANCE ASSESSMENT

This section reports the experimental setup and performance assessment for the proposed double DL-based event data coding and classification framework, considering the event data classification pipelines presented in Fig. 3. The impact on event reconstruction and classification using a PC-based representation is assessed for lossless and lossy scenarios, including voxelization, as well as conventional and DL-based PCC, using the previously selected codecs.

### A. EXPERIMENTAL SETUP

The N-Caltech101 event dataset [34], which was described in detail in Section III, was chosen in this paper since it is the most commonly used dataset for event data classification. The test split of the N-Caltech101 event dataset, composed of 1,741 event sequences, was used in the experiments for all the considered event data classification pipelines, i.e., original, voxelized, and decompressed domains.

Given that this paper converts event data to PCs through a novel Event to PC Conversion module to leverage existing PCC solutions, it is reasonable to adapt PCC quality metrics to assess the RD performance for event data. As a result, two event data quality metrics, inspired by previous PC quality metrics used for PCC, are proposed alongside the rate value to evaluate the RD performance for event data coding:

- **Rate** – Corresponds to the rate generated when compressing the input event sequence data, measured in bits-per-event (bpe), as the sum of rates for the POS and NEG polarities bitstreams, divided by the number of events in the corresponding original event sequence.

- **Peak Signal-to-Noise Ratio Event-to-Event Distance (PSNR E2E)** – Measures the event data quality based on the weighted average squared Euclidean distance, by first computing the mean squared error (MSE) between each point/event in the reference event sequence and the closest point/event with the same polarity in the decompressed event sequence. Then, the MSE is computed using the distances in the other direction, i.e., from the decompressed event sequence to the reference event sequence, and finally the PSNR is computed considering the maximum of the two MSEs. The PSNR E2E metric is inspired by the so-called PSNR Point-to-Point Distance (D1) metric, which is one of the most commonly used PC geometry quality metrics which is recommended by both the MPEG Common Test Conditions (CTC) [41] and JPEG CTTC [40].

- **Peak Signal-to-Noise Ratio Event-to-Distribution Distance (PSNR E2D)** – Measures the event data quality based on the distance between each point/event in the reference event sequence and the distribution of the closest 31 points/events with the same polarity in the decompressed event sequence, using the Mahalanobis distance as proposed in [42] for PC quality. Similar to PSNR E2E, PSNR E2D is a symmetric metric which uses the maximum of the errors between the two directions (reference to decompressed and decompressed to reference) as the final error to compute the PSNR. Contrary to the PSNR E2E quality metric, which considers distances between single points/events, PSNR E2D considers distances between points/events and distributions of points/events.

In order the results for the various processing and coding options, as well as TSF values, are comparable, a common reference for the PSNR computations has to be used. In this paper, the reference for both PNSR E2E and PSNR E2D is the original event data sequence scaled with TSF=256 (without voxelization).

Regarding the classification performance, the most widely classification metric adopted in literature is used, namely:

- **Top-k** – Represents the percentage of test examples in which the ground truth class label is in the set of $k$

prediction class labels with the highest probabilities. For this paper, $k$ is selected to be 1 (Top-1 metric) and 5 (Top-5 metric).

For comparing compression and classification performance for alternative solutions, the BD-Rate metric is used:

- **BD-Rate** – Allows comparing two performance versus rate curves, e.g., RD curves for compression, and represents the average variation in rate for the same performance, e.g., quality or classification, in percentage; a negative value indicates a rate reduction for a constant performance level, thus corresponding to 'better' solutions [43].

In the event data coding experiments, the following codecs will be relevant:

- **G-PCC lossless** – Corresponds to the G-PCC Octree standard, in lossless compression mode, using the TMC13 reference software, version v21; for event PC data coding, several TSF values are used.
- **G-PCC lossy** – Corresponds to the G-PCC Octree standard, used in lossy compression mode, using the TMC13 reference software, version v21; for event PC data coding, several TSF values and the NN method for different polarity duplicates processing are used.
- **JPEG PCC** – Corresponds to the DL-based JPEG PCC standard for geometry coding using the VM software version 3.0 (without any retraining), with spatial sampling factor 1 and block size 128 as coding parameters; for event PC data coding, the same TSF values used for G-PCC are applied and both the NN and Prob methods for different polarity duplicates processing are used.

Since the EST classifier trained model was not publicly available, the EST event data classifier model has been trained using the original N-Caltech101 event dataset training and validation splits. This involved 30 epochs, a batch size of 4, and the Adam optimizer with an initial learning rate of $10^{-4}$, reduced by a factor of two every 10,000 iterations. The training process adopted the cross-entropy as the loss function between the predicted and ground truth labels. This trained EST classifier model with original data is used for all defined classification pipelines without further retraining.

### B. ORIGINAL AND VOXELIZED EVENT DATA CLASSIFICATION PERFORMANCE

This subsection reports and analyzes the classification performance for the selected EST classifier when considering the original and voxelized event data (i.e., without coding). The classification performance for the original events will serve as the benchmark/reference for all classification pipelines, while the classification performance for the voxelized event data allows to assess the impact in terms of classification performance of the Event to PC Conversion module, notably when using different TSF values.

The TSF is a crucial parameter since it impacts both the compression and classification performances (assessed in a later subsection). While smaller TSF values generate a coarser voxelization effect, leading to a larger degradation of the event data since more points/events are discarded during the duplicate removal stage, higher TSF values may degrade the compression performance due to the larger number of events to be coded and the resulting sparsity. As such, it is important to find the TSF value offering a good trade-off between the compression and classification performances which may depend on the dataset statistical characteristics.

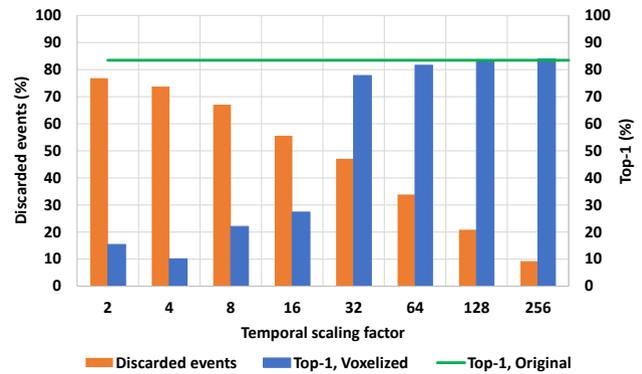

**FIGURE 4.** Impact of voxelization for different TSF values, in terms of Discarded Events (%) and Top-1 (%) classification performance. For voxelized classification performance, NN is used for different polarity duplicates processing.

Fig. 4 shows the impact of using various TSF values in the Event to PC Conversion module on the classification performance, notably in terms of the number of discarded events (%) due to voxelization and the Top-1 classification performance, when no coding is involved. The results in Fig. 4 highlight:

- The Top-1 classification performance for the original event data using the EST classifier, represented by the horizontal green line, is 83.52%; this is the benchmark/reference performance, since it was obtained with original event data.
- Lower TSF values result in a substantial increase on the percentage of discarded events, represented by the orange bars, negatively impacting the Top-1 classification performance.
- Increasing TSF rapidly decreases the percentage of discarded events and improves the Top-1 classification performance.

Fig. 4 shows that for TSF values of 64, 128, and 256, the classification performance is very close to the Original Event Data Classification performance using the EST event classifier. Thus, these TSF values are selected for further experiments involving the Decompressed Event Data Classification pipeline, described in the following subsections.

### C. EVENT DATA COMPRESSION PERFORMANCE

This subsection reports and analyzes the RD performance for coding the test split of the N-Caltech101 event dataset, using the conventional G-PCC and the DL-based JPEG PCC codecs, in the context of the Decompressed Event Data Classification pipeline. Fig. 5 shows the G-PCC lossy and JPEG PCC RD performance, considering the previously selected TSF values of 64, 128, and 256. For JPEG PCC, the results for both duplicate processing methods are presented (Nearest Neighbor-based, marked as NN, and Probabilities-based, marked as Prob).

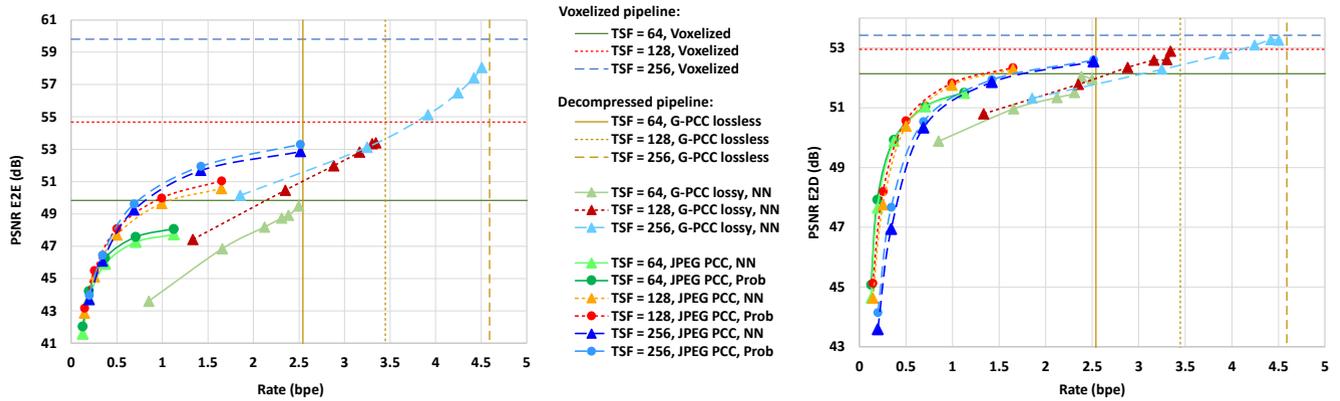

The G-PCC lossless rates are included in Fig. 5 (as vertical lines), since it would be meaningless to assess lossy compression at rates higher than the G-PCC lossless rates. The PSNR E2E and PSNR E2D values for the voxelized events using TSF equal to 64, 128 and 256, without coding, are also represented in Fig. 5. These horizontal lines mark the maximum possible reconstruction quality, which corresponds to the distortion introduced by the voxelization stage only (i.e., without compression artifacts), in the context of the Voxelized Event Data Classification pipeline. The results in Fig.5 highlight:

- JPEG PCC has a superior RD performance than G-PCC lossy for coding event data, for both PSNR E2E and PSNR E2D, especially for low to medium rates. However, for higher rates, the G-PCC lossy performance is better, being able to achieve quality levels closer to the maximum achievable quality (corresponding to the quality of the voxelized events, without coding).

- Increasing TSF consistently improves the quality (for PSNR E2E and PSNR E2D) both for the non-compressed and compressed event data. The gains in quality are related to the reduction in the number of duplicated events removed during the Event to PC Conversion when TSF is increased. Nevertheless, this effect comes at the cost of an increased rate, which can be observed for both lossless and lossy G-PCC, as well as for JPEG PCC.

- Increasing TSF always results in better PSNR E2E compression performance. However, for PSNR E2D, the gains in quality are not fully compensated by the additional rate, resulting in an inferior RD performance when TSF=256 is used. This is related to the fact that PSNR E2D is an event-to-distribution distance-based metric, which is not affected by individual point errors in the same way as is the event-to-event distance-based PSNR E2E metric. Since the PSNR E2D quality is not as severely affected by noise and the missing points resulting from voxelization and duplicate removal as PSNR E2E, the PSNR E2D quality for the several TSF values is not as significantly impacted.

- The RD performance for the two processing duplicates methods used with JPEG PCC, namely NN and Prob, shows that the Prob method is more effective in handling duplicates with different polarities, especially at lower bitrates. In fact, Fig. 5 shows that: i) for PSNR E2E, the Prob method provides better RD performance than the NN method across all rates; ii) for PSNR E2D, the RD advantage of the Prob method is less noticeable since PSNR E2E is more sensitive to individual point positions than PSNR E2D.

The RD performance results show that the compression artifacts generated by lossy event data coding using G-PCC lossy and JPEG PCC negatively affect the reconstruction quality of the decompressed event data although this impact may be controlled and made as low as appropriate. In this context, it is important to assess the impact of coding on the event processing tasks performance, notably event data classification.

### D. LOSSY DECOMPRESSED EVENT DATA CLASSIFICATION PERFORMANCE

This subsection reports and analyzes the classification performance when G-PCC lossy and JPEG PCC decompressed event data are classified using the EST classifier, notably in comparison with the original and voxelized classification performances. As mentioned before, the classification performance is evaluated using the Top-k classification metric for the most commonly used $k$ values, i.e., 1 and 5.

The classification performance of all defined event data classification pipelines, incorporating the different polarity duplicate processing methods outlined in Subsection VI.B, is depicted in Fig. 6. This figure presents the average classification performance as a function of the rate (in bpe) for the Top-1 and Top-5 classification metrics for the test split of the N-Caltech101 event dataset. The EST classifier, trained on the original event data (training and validation splits of the N-Caltech101 event dataset), is applied without retraining for all the tested event data classification pipelines. The results from Fig. 6 highlight:

- For the Voxelized Event Data Classification pipeline, the classification performance for voxelized data is similar to the classification performance for the original event data, for both Top-1 and Top-5 classification metrics, for the tested TSF values (as also shown in Fig. 4).
- As expected, for the Decompressed Event Data Classification pipeline, there is an improvement in both the Top-1 and Top-5 classification performances when the rate increases.
- It is important to note that the classification performance for decompressed events using the JPEG PCC and G-PCC lossy closely approaches the classification performance for the original events, even for rates well below those required for G-PCC lossless compression.
- Regarding the comparison between JPEG PCC and G-PCC in terms of classification performance, JPEG PCC decompressed events offer better classification performance than G-PCC lossy decompressed events at similar rates, for both Top-1 and Top-5. This suggests that the artifacts generated by JPEG PCC are not as impactful as the G-PCC compression artifacts and still enable the decompressed signal to preserve important features necessary for accurate event data classification. At the lowest rates, both JPEG PCC and G-PCC lossy result in severe losses in classification performance. However, at the lowest rates both JPEG PCC and G-PCC lossy introduce compression artifacts that severely impact the associated classification performance. Nevertheless, these rates may be less relevant since most applications do not accept such poor quality (decompressed) event data.
- For JPEG PCC decompressed events, the EST

classifier performs better in classifying events when the Prob duplicate removal method is used, in alternative to the NN method, especially for low rates.

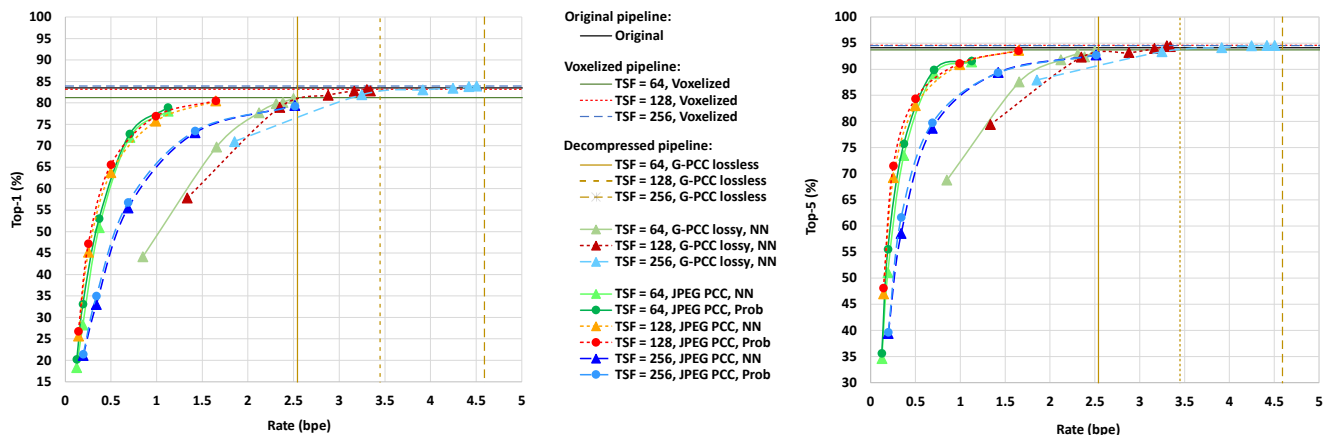

**FIGURE 6.** Top-1 (left) and Top-5 (right) classification performance for the EST event data classifier using various TSF values and duplicates processing methods.

To provide more information about the relative classification performance of G-PCC lossy and JPEG PCC, Table I presents BD-Rate values, for Top-1 and Top-5, for the lossy decompressed scenarios also considered in Fig. 6. This comparison is made assuming as reference the classification performance for G-PCC lossy with TSF=64. The results in Table I highlight:

- The Top-1 and Top-5 BD-Rate values show gains, i.e., negative values and thus rate reductions, for every JPEG PCC lossy coding option, while the BD-Rate values for G-PCC lossy using TSF values different from 64 show a decrease in classification performance, i.e., positive values, and thus rate increases for the same classification performance.
- For JPEG PCC decompressed events, the BD-Rate values indicate rate savings of 65.45% and 65.25% for Top-1 and Top-5, respectively, for the Prob method compared to the NN method, for TSF=128. For TSF=128, the BD-Rate gains are larger than for the other TSF values, for both the Prob and NN methods, although good gains are also obtained for TSF=64.

**TABLE I**
**BD-RATE PERFORMANCE FOR DECOMPRESSED EVENT DATA CLASSIFICATION PIPELINE FOR VARIOUS TSF VALUES WHEN CODING WITH G-PCC LOSSY AND JPEG PCC, ADOPTING AS REFERENCE G-PCC LOSSY WITH TSF=64, NN**

| PCC | TSF, different polarity processing method | BD-Rate | |
|---|---|---|---|
| | | Top-1 | Top-5 |
| G-PCC lossy | TSF = 128, NN | +2.57% | +4.43% |
| | TSF = 256, NN | +8.76% | +15.92% |
| JPEG PCC | TSF = 64, NN | -61.30% | -61.20% |
| | TSF = 64, Prob | -63.01% | -63.87% |
| | TSF = 128, NN | -62.63% | -62.49% |
| | TSF = 128, Prob | **-65.45%** | **-65.25%** |
| | TSF = 256, NN | -31.95% | -34.41% |
| | TSF = 256, Prob | -34.53% | -36.96% |

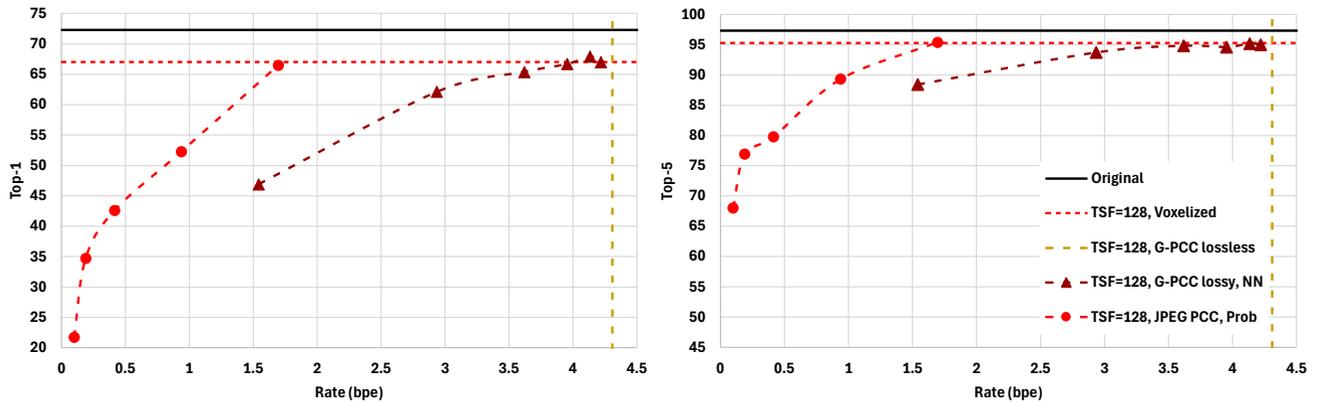

**Figure 7.** Top-1 (left) and Top-5 (right) classification performance for the CIFAR10-DVS dataset test split using the EST classifier with TSF=128. G-PCC lossy and JPEG PCC use NN and Prob to perform the different duplicates processing, respectively.

Although currently most event data coding works primarily focus on lossless coding, notably using G-PCC lossless, these results show that the usage of lossy coding, notably using JPEG PCC, allows to achieve comparable classification performances for the EST event data classifier for rates which are much lower than the G-PCC lossless rate. This effectively opens the doors for the adoption of lossy coding of event data.

### E. GENERALIZATION CAPABILITIES

This subsection presents an analysis of the classification performance on an additional event dataset, to evaluate the generalization capabilities of the proposed solution. For this purpose, the CIFAR10-DVS event dataset is selected, which is an event dataset based on the CIFAR-10 image dataset, originally produced by the Canadian Institute for Advanced Research [35].

The CIFAR10-DVS dataset is designed for classification tasks and based on the 10 classes version of the CIFAR image dataset, which includes 60,000 color images of resolution 32×32. It is worthwhile to mention that the acquisition process for CIFAR10-DVS differs from that of the N-Caltech101 event dataset, as detailed in Section III. In the CIFAR10-DVS dataset, the images are moving while the event camera remains stable, which is not the case for the N-Caltech101 dataset. For the CIFAR10-DVS dataset, 1,000 images from each class are selected, up-sampled to 512×512 and converted into event sequences [35]. This conversion is achieved by displaying and moving the images on a screen in front of a dynamic vision sensor (DVS) with a resolution of 128×128. As a result, the CIFAR10-DVS dataset consists of 10,000 event sequences, organized into 10 classes corresponding to the original image CIFAR-10 classes, e.g. airplane, bird, cat, etc. Each class contains 1,000 event sequences, each lasting 1 second. Using the CIFAR10-DVS dataset with the proposed coding and classification solution required the following

processing steps:

1. Following the protocol outlined in [45], the CIFAR10-DVS dataset is randomly split into training and testing datasets using a 9:1 ratio.

2. The EST classifier (previously trained for the N-Caltech101 event dataset) is retrained using the original CIFAR10-DVS training split, since the type and number of classes for the two datasets are very different. During training, the defined training split was further randomly divided into an 80% training portion and a 20% validation portion, applied independently to each of the 10 labeling classes.

3. The events from the CIFAR10-DVS dataset test split are processed according to the same procedure previously used for the N-Caltech101 test dataset: two PCs are generated from each event sequence of the CIFAR10-DVS dataset test split (one per polarity), using the proposed Event to PC Conversion tool and the good performing value TSF=128, previously determined for the test split of the N-Caltech101 dataset.

4. The CIFAR10-DVS PCs are encoded using the selected PCC solutions, notably JPEG PCC, G-PCC lossy and G-PCC lossless. After decoding, the EST classifier is used to classify the decompressed events for all codecs.

The results in Fig. 7 demonstrate that the comparative performance of the selected solutions for the CIFAR10-DVS dataset test split is similar to the comparative performance previously obtained for the N-Caltech101 dataset test split, notably confirming the main performance claims, notably:

- For higher rates, the classification performance of the decompressed events closely approaches the classification performance of the original events. For JPEG PCC, this happens for rates well below those required for G-PCC lossless coding. This confirms the previous claim that it is possible to use lossy coding of event data, notably with DL-based JPEG PCC, while achieving a classification performance comparable to that of the original event data at significantly lower rates.

- The classification performance for JPEG PCC decompressed events is higher than the classification performance achieved for G-PCC lossy decompressed events. This confirms the previous claim that coding with a DL-based codec like JPEG PCC is more effective than coding with a state-of-the-art conventional codec like G-PCC lossy.

These conclusions demonstrate that the performance claims made for the N-Caltech101 dataset generalize to another event dataset, notably the CIFAR10-DVS dataset, in this case using the same TSF value. However, it is important to observe that, for other event datasets, with different characteristics, it is possible that the optimum compression and classification performances may be achieved for another TSF value.

### F. IMPROVEMENT PERSPECTIVES

This section discusses the improvement perspectives for more efficient event data coding and classification. The results in Fig. 6 show that the classification performance drops for lower coding rates, thus limiting the rate savings for a target classification performance; this effect is expected in the same way for any lossy codec, be it conventional or DL-based. However, DL-based codecs offer a major advantage over their conventional counterparts since this limitation may be addressed by performing the CV task at hand, in this case classification, in the compressed domain, this means directly processing the compressed stream without decoding.

This target has been adopted in the JPEG AI and JPEG Pleno Point Cloud Coding standards' scopes and has already been demonstrated in the literature for images [46], [28] and PCs [26],[27]. Their scopes aim to provide media representations that are not only able to deliver efficient compression for human visualization but also effective performance for CV tasks in the compressed domain. This is a novel research area not yet explored in the event data related research, made possible due to the advances in DL-based data processing.

The compressed domain processing approach offers two key benefits: firstly, it enhances the CV task performance, particularly at lower rates, as DL-based representation features are directly extracted from the original event data, avoiding the compression artifacts introduced during decoding; secondly, it simplifies the process by directly using the compressed representation for CV processing, eliminating the need for decoding and subsequent feature extraction.

Previous research conducted by the authors of this paper has achieved significant improvements in compressed domain PC classification compared to original, voxelized and decompressed domain classification performances, especially at lower coding rates [26], [27]. It is expected that the same will happen in the future for DL-based event data coding.

### VIII. CONCLUSIONS AND FUTURE WORK

Event cameras capture asynchronous streams of per-pixel brightness changes, known as "events", which can be represented using multiple modalities, notably as PC data. This paper proposes the first double DL-based solution for efficient event data coding and classification and assesses the impact of lossy DL-based PCC using the JPEG PCC standard on DL-based event data classification, notably using the EST classifier. The proposed solution includes novel dedicated modules for converting event data to event PC data and vice versa. The results highlight that the EST classification performance with JPEG PCC decompressed events may reach the reference classification performance for appropriate rates, below the G-PCC lossless rates and also below the G-PCC lossy rates, meaning that it is possible and beneficial to adopt JPEG PCC (lossy) event data coding before CV processing when there are limited transmission and storage resources.

Currently, event-based cameras do not incorporate any efficient coding techniques in their data output process. However, in the future, these cameras may perform real-time lossy coding and output lossy compressed event data, notably if a low complexity encoder is available. This paper demonstrates that, in such scenarios, the use of DL-based PCC has advantages in terms of RD performance and enables future compressed domain CV tasks. This advancement could significantly enhance the capabilities of event-based cameras, making them more versatile and effective for a wider range of applications, notably involving CV tasks.

Future work will focus on developing a JPEG PCC compressed domain event data classifier to mitigate the impact of compression artifacts, thus reducing the rate required for a given target classification performance; this may be achieved while simultaneously reducing the overall system complexity. Naturally, other CV tasks may be also addressed, such as object detection and segmentation, along with performance assessment on appropriate datasets.

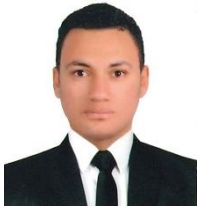

**ABDELRAHMAN SELEEM** (Graduate Student Member, IEEE) received the B.Sc. and M.Sc. degree in computer science from the Computer Science-Mathematics Department, Faculty of Science, South Valley University, Egypt in 2016 and 2021, respectively. From March 2022 to the present, he is a Ph.D. student at the Department of Electrical and Computer Engineering, Instituto Superior Técnico, University of Lisbon, Lisbon, Portugal, and a junior researcher at Instituto de Telecomunicações, Portugal. From April 2017 to August 2020, he worked as a Teaching Assistant at the Computer Science-Mathematics Department, Faculty of Science, South Valley University, Qena, Egypt. He worked as a Teaching Assistant and Lecturer Assistant from September 2020 to May 2021 and from June 2021 to February 2022, respectively, at the Computer Science Department, Faculty of Computers and Information, South Valley University, Qena, Egypt. He is actively contributing to the standardization efforts of JPEG on learning-based point cloud coding. His research interests include deep learning, computer vision, image processing, point cloud coding and analysis.

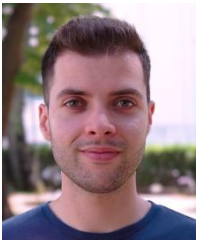

**ANDRÉ F. R. GUARDA** (Member, IEEE) received his B.Sc. and M.Sc. degrees in Electrotechnical Engineering from Instituto Politécnico de Leiria, Portugal, in 2013 and 2016, respectively, and the Ph.D. degree in Electrical and Computer Engineering from Instituto Superior Técnico, Universidade de Lisboa, Portugal, in 2021. He has been a researcher at Instituto de Telecomunicações since 2011, where he currently holds a Post-Doctoral position. His main research interests include multimedia signal processing and coding, with particular focus on point cloud coding with deep learning. He has authored several publications in top conferences and journals in this field, and is actively contributing to the standardization efforts of JPEG and MPEG on learning-based point cloud coding.

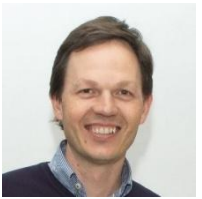

**NUNO M. M. RODRIGUES** (Senior Member) graduated in electrical engineering in 1997, received the M.Sc. degree from the Universidade de Coimbra, Portugal, in 2000, and the Ph.D. degree from the Universidade de Coimbra, Portugal, in 2009, in collaboration with the Universidade Federal do Rio de Janeiro, Brazil. He is a Professor in the Department of Electrical Engineering, in the School of Technology and Management of the Polytechnic University of Leiria, Portugal and a Senior Researcher in Instituto de Telecomunicações, Portugal. He has coordinated and participated as a researcher in various national and international funded projects. He has supervised three concluded PhD theses and several MSc theses. He is co-author of a book and more than 100 publications, including book chapters and papers in national and international journals and conferences. His research interests include several topics related with image and video coding and processing, for different signal modalities and applications. His current research is focused on deep learning-based techniques for point cloud coding and processing.

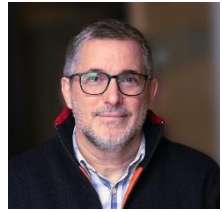

**FERNANDO PEREIRA** (Fellow, IEEE) graduated in electrical and computer engineering in 1985 and received the M.Sc. and Ph.D. degrees in 1988 and 1991, respectively, from Instituto Superior Técnico, Technical University of Lisbon. He is currently with the Department of Electrical and Computer Engineering, Instituto Superior Técnico and Instituto de Telecomunicações, Lisbon, Portugal. He is also the JPEG Requirements Subgroup Chair. Recently, he has been one of the key designers of the JPEG Pleno and JPEG AI standardization projects. He has contributed more than 300 papers in international journals, conferences, and workshops, and made several tens of invited talks and tutorials at conferences and workshops. His research interests include visual analysis, representation, coding, description and adaptation, and advanced multimedia services. He was an IEEE Distinguished Lecturer, in 2005, and elected as a fellow of IEEE in 2008 for "Contributions to object-based digital video representation technologies and standards." Since 2013, he has been a EURASIP Fellow for "Contributions to digital video representation technologies and standards." Since 2015, he has been a fellow of IET. He is or has been a member of the Editorial Board of the Signal Processing Magazine, an Associate Editor of IEEE TRANSACTIONS ON CIRCUITS AND SYSTEMS FOR VIDEO TECHNOLOGY, IEEE TRANSACTIONS ON IMAGE PROCESSING, IEEE TRANSACTIONS ON MULTIMEDIA, and IEEE Signal Processing Magazine. He has been elected to serve on the Signal Processing Society Board of Governors in the Capacity of Member-at-Large (2012) and (2014–2016). He has been the Vice President of the IEEE Signal Processing Society, from 2018 to 2020. He has also been elected to serve on the European Signal Processing Society Board of Directors (2015–2018). He is an Area Editor of the Signal Processing: Image Communication journal and an Associate Editor of the EURASIP Journal on Image and Video Processing. From 2013 to 2015, he was the Editor-in-Chief of the IEEE JOURNAL OF SELECTED TOPICS IN SIGNAL PROCESSING.